\pdfoutput=1
\documentclass[twocolumn]{article}

\usepackage{wacv}

\usepackage[pagebackref,breaklinks,colorlinks]{hyperref}

\usepackage{enumitem}
\usepackage{tikz}
\usepackage{pgfplots}
\usepackage{microtype}
\usepackage{xcolor}
\usepackage{bbold}
\usepackage{mathtools}
\usepackage{ifthen}

\setitemize{noitemsep,parsep=0pt,partopsep=0pt,leftmargin=*}

\usepgfplotslibrary{fillbetween}
\usetikzlibrary{patterns}
\usetikzlibrary{positioning}

\pgfplotsset{
    box plot/.style={
        /pgfplots/.cd,
        only marks,
        mark=-,
        mark size=1em,
        /pgfplots/error bars/.cd,
        y dir=plus,
        y explicit,
    },
    box plot box/.style={
        /pgfplots/error bars/draw error bar/.code 2 args={%
            \draw  ##1 -- ++(1em,0pt) |- ##2 -- ++(-1em,0pt) |- ##1 -- cycle;
        },
        /pgfplots/table/.cd,
        y index=2,
        y error expr={\thisrowno{3}-\thisrowno{2}},
        /pgfplots/box plot
    },
    box plot top whisker/.style={
        /pgfplots/error bars/draw error bar/.code 2 args={%
            \pgfkeysgetvalue{/pgfplots/error bars/error mark}%
            {\pgfplotserrorbarsmark}%
            \pgfkeysgetvalue{/pgfplots/error bars/error mark options}%
            {\pgfplotserrorbarsmarkopts}%
            \path ##1 -- ##2;
        },
        /pgfplots/table/.cd,
        y index=4,
        y error expr={\thisrowno{2}-\thisrowno{4}},
        /pgfplots/box plot
    },
    box plot bottom whisker/.style={
        /pgfplots/error bars/draw error bar/.code 2 args={%
            \pgfkeysgetvalue{/pgfplots/error bars/error mark}%
            {\pgfplotserrorbarsmark}%
            \pgfkeysgetvalue{/pgfplots/error bars/error mark options}%
            {\pgfplotserrorbarsmarkopts}%
            \path ##1 -- ##2;
        },
        /pgfplots/table/.cd,
        y index=5,
        y error expr={\thisrowno{3}-\thisrowno{5}},
        /pgfplots/box plot
    },
    box plot median/.style={
        /pgfplots/box plot
    }
}

\definecolor{sheetraw}{RGB}{78,123,38}
\definecolor{sheetdelta}{RGB}{180,138,36}
\definecolor{sheetdeltaall}{RGB}{31,120,180}

\newcommand\sheetrawlow{92}
\newcommand\sheetrawmult{50}
\newcommand\sheetdeltamult{80}
\newcommand{\sheettable}[9]{
\begin{tikzpicture}[scale=0.8, inner sep=0.]
    \foreach \y [count=\n] in {#1}{
        \foreach \x [count=\m] in \y {
            \pgfmathsetmacro \clr {(\x-\sheetrawlow)*\sheetrawmult}
            \node[fill=sheetraw!\clr, minimum size=8mm] at (\m,-\n) {\scriptsize\x};
        }
    }
    \foreach \dely [count=\n] in {#2}{
        \pgfmathsetmacro \delyloc {#7+1}
        \pgfmathsetmacro \delyclr {\dely*\sheetdeltamult}
        \node[fill=sheetdelta!\delyclr, minimum size=8mm] at (\delyloc,-\n) {\scriptsize\dely};
    }
    \foreach \delx [count=\m] in {#3}{
        \pgfmathsetmacro \delxclr {\delx*\sheetdeltamult}
        \node[fill=sheetdelta!\delxclr, minimum size=8mm] at (\m,-6) {\scriptsize\delx};
    }
    \pgfmathsetmacro \delclr {#4*\sheetdeltamult}
    \pgfmathsetmacro \delyloc {#7+1}
    \pgfmathsetmacro \lblloc {(#7+2)/2}
    \node[] at (\lblloc,0.6) {\textbf{\footnotesize#6}};
    \node[] at (\delyloc,0) {\textbf{\footnotesize$\Delta_{max}$}};
    \foreach \xlabel [count=\m] in {#8}{
        \node[minimum size=8mm] at (\m,0) {\scriptsize\xlabel};
    }
    \ifthenelse{#9>0}{
    \node[rotate=90] at (0,-3) {\textbf{\footnotesize#5}};
    \node[rotate=90] at (0,-6) {\textbf{\footnotesize$\Delta_{max}$}};
    }{}
\end{tikzpicture}}
\newcommand{\sheetheatmap}[3]{
\begin{tikzpicture}
\scriptsize
\begin{axis}[
    hide axis,
    scale only axis,
    height=0pt,
    width=0pt,
    colormap={}{color=(white) color=(#3)},
    colorbar horizontal,
    point meta min=#1,
    point meta max=#2,
    colorbar style={
        width=1.6cm,
        xtick={#1,#2},
    }]
    \addplot [draw=none] coordinates {(0,0)};
\end{axis}
\end{tikzpicture}}

\begin{document}

\title{An Empirical Investigation into Benchmarking Model Multiplicity for Trustworthy Machine Learning: A Case Study on Image Classification}

\author{Prakhar Ganesh\\
Mila, Quebec AI Insitute\\
{\tt\small prakhar.ganesh@mila.quebec}
}

\maketitle
 
\begin{abstract}
Deep learning models have proven to be highly successful. Yet, their over-parameterization gives rise to model multiplicity, a phenomenon in which multiple models achieve similar performance but exhibit distinct underlying behaviours. This multiplicity presents a significant challenge and necessitates additional specifications in model selection to prevent unexpe   cted failures during deployment. While prior studies have examined these concerns, they focus on individual metrics in isolation, making it difficult to obtain a comprehensive view of multiplicity in trustworthy machine learning. Our work stands out by offering a one-stop empirical benchmark of multiplicity across various dimensions of model design and its impact on a diverse set of trustworthy metrics.

In this work, we establish a consistent language for studying model multiplicity by translating several trustworthy metrics into accuracy under appropriate interventions. We also develop a framework, which we call multiplicity sheets, to benchmark multiplicity in various scenarios. We demonstrate the advantages of our setup through a case study in image classification and provide actionable insights into the impact and trends of different hyperparameters on model multiplicity. Finally, we show that multiplicity persists in deep learning models even after enforcing additional specifications during model selection, highlighting the severity of over-parameterization. The concerns of under-specification thus remain, and we seek to promote a more comprehensive discussion of multiplicity in trustworthy machine learning.
\end{abstract}

\section{Introduction}
\label{sec:introduction}

Deep learning has experienced a remarkable rise in recent years \cite{sharifani2023machine,ramesh2021zero,rombach2022high}, with highly sophisticated and over-parameterized models leading the way \cite{dehghani2023scaling,scao2022bloom}. Consequently, these cutting-edge models find application across a diverse set of domains, including image processing \cite{voulodimos2018deep}, natural language \cite{lin2022survey}, healthcare \cite{esteva2019guide}, finance \cite{ahmed2022artificial}, judiciary systems \cite{yassine2023using}, and more, showcasing their versatility and potential impact. However, the increasing deployment of these models has sparked concerns about their trustworthiness. To confront these issues head-on, the global community has embraced a range of trustworthy machine learning practices and metrics \cite{kearns2019ethical,varshney2019trustworthy}. These efforts are geared towards ensuring that these models are not only accurate in their predictions, but are also fair to various groups in the dataset \cite{barocas2017fairness,mehrabi2021survey}, robust to distribution shifts \cite{subbaswamy2021evaluating}, maintain the privacy of the individuals whose data was collected \cite{dwork2014algorithmic}, and secure against adversarial attacks \cite{chakraborty2021survey}. These metrics collectively strive to make deep learning deployments more reliable, fostering trust and acceptance in its widespread applications.

Alongside the discussion of trustworthy ML, the presence of multiplicity in deep learning has emerged as a significant concern yet a welcome opportunity \cite{black2022model,d2022underspecification}. Model multiplicity is the existence of multiple high-performing models that achieve similar accuracy on a given task but can display diverse predictive behaviours due to varying decision boundaries and underlying learned functions. Model multiplicity is the result of an under-specified and over-parameterized training regime, and can be affected by design choices like model architectures, hyperparameters, training configurations, or even arbitrary choices like the randomness in training.
% Consequently, they may provide comparable accuracies but with varying decision boundaries and underlying learned functions.

Model multiplicity in deep learning has significant implications. For instance, it has been shown that changes in the training configuration can lead to considerable variations in the biases present in a model \cite{ganesh2023impact,perrone2021fair}. Deploying such models without considering the impact of multiplicity can result in the unintentional deployment of unfair models in real-world applications. Conversely, if we manage multiplicity with appropriate constraints, it presents an opportunity to deploy fairer models without compromising its utility. Thus, addressing the challenges of model multiplicity is a crucial step towards creating trustworthy systems.

Existing literature on investigating model multiplicity is limited to specialized settings that do not generalize. For instance, Somepalli \etal~\cite{somepalli2022can} provides an empirical quantification of similarity in the decision boundary of two models. However, the similarity of decision boundaries may not necessarily provide any information about its trustworthiness. Models with significantly different decision boundaries can still provide similar accuracy, fairness, robustness, security, and privacy. Similarly, Ganesh \etal~\cite{ganesh2023impact} investigates the impact of random seeds on fairness, but their discussion focuses on model predictions, and thus may not extend to other trustworthy metrics like robustness, security, or privacy. Furthermore, these investigations are not directly comparable to each other. For example, a $70\%$ agreement between the decision boundaries of two models as defined by Somepalli \etal~\cite{somepalli2022can} has no comparative value to a $10\%$ gap in equalized odds (a fairness metric) between the same set of models \cite{ganesh2023impact}. Thus, while these works provide a deeper investigation into a single metric in isolation, they fail to provide a comprehensive view of the overall trends of multiplicity.

In this paper, we address this gap by proposing a framework to measure multiplicity that can not only dive deeper into multiplicity trends for a single metric but also provide comparisons across different metrics. We start by converting various trustworthy metrics to a common scale, which we refer to as \textit{accuracy under intervention}, facilitating the comparison of multiplicity across different metrics. We then create \textit{multiplicity sheets} that capture the multiplicity of accuracy under intervention for each metric separately. 
% These sheets enable readers to study the multiplicity trends for a single metric. 
To illustrate our framework, we present an image classification case study that compares multiplicity across model hyperparameters, random seeds, and architecture choices, repeating the setup for various trustworthy ML metrics, namely fairness, robustness, privacy, and security.

We end our discussion by presenting the results of combining various metrics together to improve the model specification and reduce multiplicity. However, despite following the recommendations of recent literature and providing additional specifications using trustworthy metrics, our study reveals that model multiplicity can still create unforeseen failure cases. This highlights the need for future research to gain a more holistic understanding of model multiplicity.

\paragraph{Setting Expectations and Our Contributions: } Before delving into our contributions, it is essential to first clarify the scope of our work. Our goal is not to present novel findings on the multiplicity of any specific metric. In fact, we will revisit many existing results in the literature during our case study. Rather, we seek to establish a normative language to record model multiplicity that can be used to highlight multiplicity trends across different metrics, thus providing an overall picture of multiplicity in deep learning.

More specifically, our contributions are:

\begin{itemize}
    \item We propose a standardized framework to measure and study model multiplicity in deep learning.
    \begin{itemize}
        \item We introduce a new class of metrics called accuracy under intervention. We showcase techniques to convert any metric into accuracy using appropriate interventions, thus providing a common scale of comparison.
        \item We suggest using multiplicity sheets, a comprehensive yet compact method to record and study model multiplicity for any target metric.
    \end{itemize}
    \item We present a case study of model multiplicity in image classification, by providing an empirical benchmark to highlight the advantages of our framework.
    \begin{itemize}
        \item We take an all-encompassing view of model multiplicity and its impact on trustworthy ML by comparing multiplicity across fairness, robustness, privacy, and security.
        \item We study the influence of various axes of model variations on multiplicity, including model architecture, training randomness, and hyperparameter choices.
    \end{itemize}
    \item We combine several trustworthy metric specifications to challenge over-parameterization and assess its impact on multiplicity. Despite this, we see persistent multiplicity on trustworthy issues not seen during model selection, underscoring the need for better safeguards against multiplicity when deploying models in the real world.
\end{itemize}

\begin{figure*}
    \centering
    \input{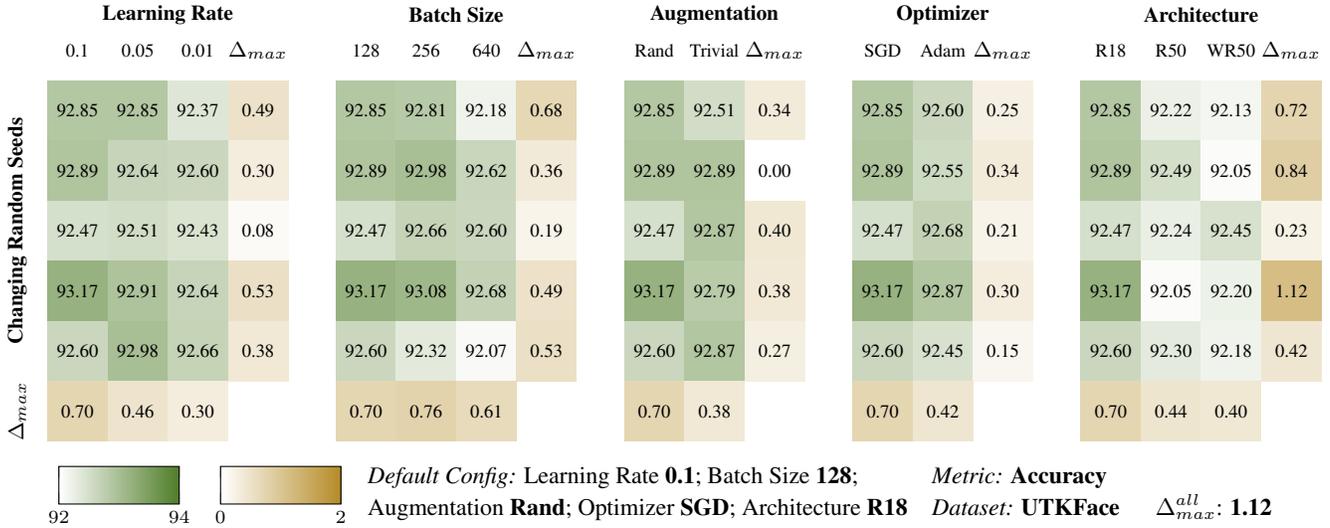}
    \caption{\textit{Multiplicity sheet} for \textbf{Accuracy} on \textbf{UTKFace} dataset. R18/50: ResNet-18/50; WR50: WideResNet-50x2. This multiplicity sheet records accuracy scores across different hyperparameter choices and random seeds, representing the first level of readability without any loss of information. It then aggregates multiplicity by measuring $\Delta_{max}$ across random seeds for each hyperparameter and across hyperparameter choices for each random seed. This is the second level of readability, vital for extracting multiplicity trends. For instance, by studying the $\Delta_{max}$ values, we see the equal importance of both random seeds and hyperparameter choices on accuracy multiplicity. Finally, we aggregate the overall multiplicity $\Delta_{max}^{all}$, i.e., the third level of readability, condensing accuracy multiplicity for UTKFace into a single value.}
    \label{fig:accuracy_multiplicity_utkface}
\end{figure*}

\section{Measuring Multiplicity}

We will start by discussing our framework to study multiplicity. We introduce the concept of \textit{accuracy under intervention} to translate any metric into an accuracy metric, followed by our proposal to use \textit{multiplicity sheets} to record and compare the said accuracy under intervention.

\subsection{Accuracy Under Intervention}

We want to establish a standardized way to measure model multiplicity that would allow easy comparison across different scenarios. However, measuring multiplicity for various trustworthy objectives relies on vastly different metrics, making it a complex task. For instance, comparing accuracy multiplicity (difference in accuracy) with security multiplicity (difference in minimum adversarial distance to flip the label), is not straightforward. These two metrics are not comparable since they are based on different factors, one on performance and the other on distance.

We need a method that can translate these metrics to a common scale for a fair comparison. To achieve this, we propose converting each metric into accuracy through appropriate interventions. Simply put, we want to measure model accuracy under a well-designed intervention that represents a proxy for our original trustworthy metric. For instance, when testing the security of a model against adversarial attacks, instead of measuring the minimum adversarial distance to flip the label, we can measure the model accuracy under a fixed adversarial distance budget. Once a metric is translated to accuracy under such an intervention, we can now compare them directly to each other and get a comprehensive understanding of the multiplicity. More details on the specific interventions for each metric are present in Section \ref{sec:utkface}.

\subsection{Multiplicity Sheets}

After converting every metric to accuracy under intervention, we propose a method for recording these values to facilitate easy comparison and visualization. We want a method that can provide both summaries for a quicker scan and detailed results for a more in-depth analysis. To achieve this, we create multiplicity sheets, a straightforward and highly intuitive approach to documenting multiplicity.

A multiplicity sheet is a collection of tables, where each table compares two axes of multiplicity. The information in our multiplicity sheets has three levels of readability. The first level shows the raw metric scores, in this case, accuracy under intervention, ensuring no loss of information. The second level aggregates multiplicity across each axis in every table by taking the maximum difference in scores denoted by $\Delta_{max}$. This allows easy visualization of various trends and the influence of different hyperparameters on model multiplicity. Finally, the third level further aggregates the complete multiplicity sheet by taking the maximum difference across all raw metric scores to get a single value representing the overall multiplicity of the given metric, denoted as $\Delta_{max}^{all}$. Given that we use accuracy under intervention for all metrics, $\Delta_{max}^{all}$ serves as a useful measure to compare multiplicity across different metrics, i.e., different multiplicity sheets.

An example of a multiplicity sheet can be seen in Fig. \ref{fig:accuracy_multiplicity_utkface}, where we record the accuracy multiplicity on the UTKFace dataset under various training configurations (more details in Section \ref{sec:utkface}). Throughout our paper, we will designate one axis of multiplicity in each table to be the random seeds. This is to filter chance trends when comparing different hyperparameters by balancing them against multiple runs with changing random seeds. It should be noted that multiplicity sheets can be created for any metric, not just accuracy under intervention. However, using accuracy under intervention allows us to compare the multiplicity trends across different sheets, which wouldn't be possible with just any metric. We will now move to our case study to highlight the benefits of our framework, while also providing an empirical benchmark of multiplicity in image classification that can be directly useful for researchers and practitioners.
\section{Image Classification on UTKFace}
\label{sec:utkface}

To demonstrate the utility of our framework, we will perform a case study of the model multiplicity in image classification on the UTKFace dataset. 
% We selected the UTKFace dataset due to the availability of sensitive attributes, which are essential for measuring fairness, and the existence of similar facial image datasets like FairFace \cite{}, to test out-of-domain robustness. 
We first outline our experimental setup, followed by a comprehensive discussion of multiplicity in fairness, robustness, privacy, and security. To provide diversity in experiments, we also perform a separate case study on the CIFAR10 dataset in Appendix \ref{sec:;app_cifar10}.

\begin{figure*}
    \centering
    \input{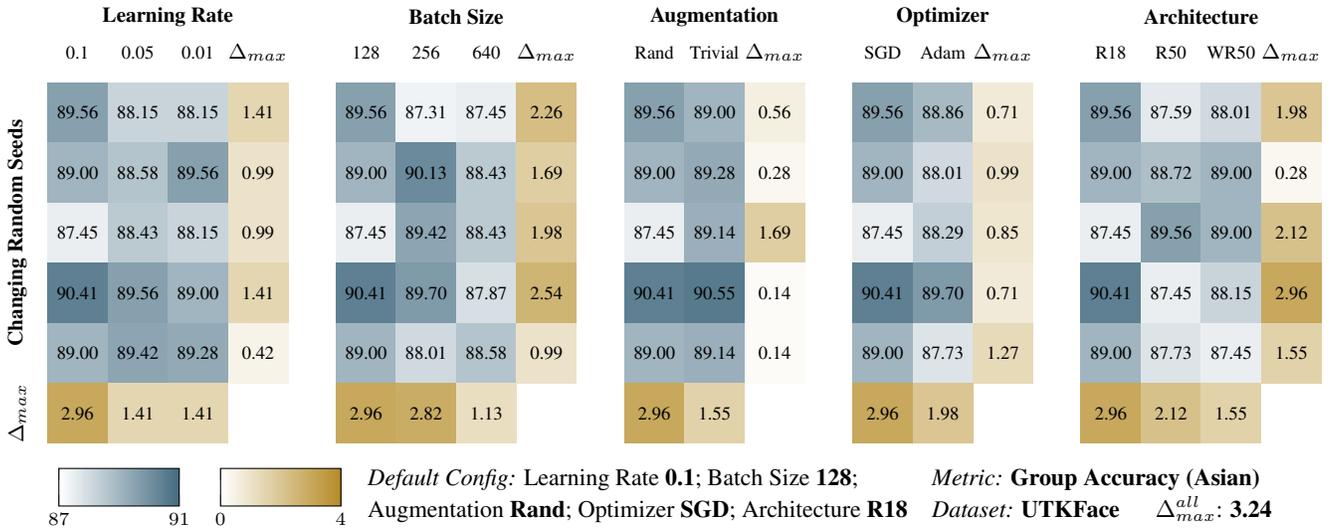}
    \caption{\textit{Multiplicity sheet} for \textbf{Group Accuracy (Asian)} on \textbf{UTKFace} dataset. R18/50: ResNet-18/50; WR50: WideResNet-50x2. Among various hyperparameter choices, the batch size and architecture stand out in their influence on fairness multiplicity. However, it is the variance across changing random seeds that overshadows all other sources of multiplicity, making it the most important factor for fairness during model selection. The overall fairness multiplicity ($\Delta_{max}^{all}$) is also almost 3 times higher than the accuracy multiplicity seen earlier.}
    \label{fig:fairness_multiplicity_utkface}
\end{figure*}

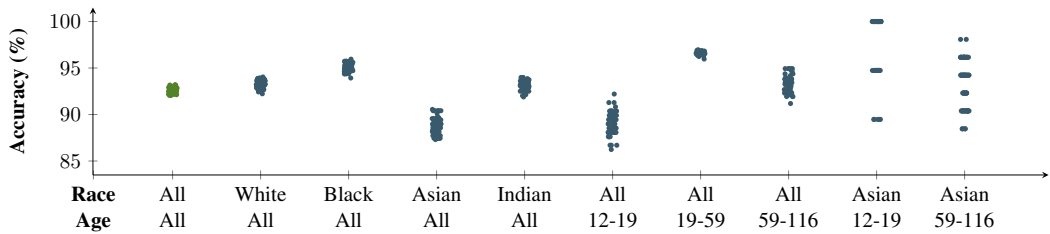
\begin{figure*}
    \centering
    \begin{tikzpicture}[scale=0.8]
    \begin{axis}[
        width=1.\textwidth,
        height=0.25\textwidth,
        xmin=1,
        ymin=85, ymax=100,
        ylabel=\textbf{Accuracy (\%)},
        axis lines=left,
        enlarge x limits = true,
        enlarge y limits = true,
        xtick={0.1,1,2,3,4,5,6,7,8,9,10},
        xticklabels={\textbf{Race}\\ \textbf{Age}, All\\ All, White\\ All, Black\\ All, Asian\\ All, Indian\\ All, All\\ 12-19, All\\ 19-59, All\\ 59-116, Asian\\ 12-19, Asian\\ 59-116},
        xticklabel style={align=center}
    ]

    \addplot[fill={rgb:red,78;green,123;blue,38}, only marks,draw opacity=0,mark size=1.2pt] table [x=x, y=y] {data/accuracy.dat};

    \addplot[fill={rgb:red,66;green,106;blue,127}, only marks,draw opacity=0,mark size=1.2pt] table [x=x, y=y] {data/accuracy_white.dat};

    \addplot[fill={rgb:red,66;green,106;blue,127}, only marks,draw opacity=0,mark size=1.2pt] table [x=x, y=y] {data/accuracy_black.dat}; 

    \addplot[fill={rgb:red,66;green,106;blue,127}, only marks,draw opacity=0,mark size=1.2pt] table [x=x, y=y] {data/accuracy_asian.dat}; 

    \addplot[fill={rgb:red,66;green,106;blue,127}, only marks,draw opacity=0,mark size=1.2pt] table [x=x, y=y] {data/accuracy_indian.dat};  

    \addplot[fill={rgb:red,66;green,106;blue,127}, only marks,draw opacity=0,mark size=1.2pt] table [x=x, y=y] {data/accuracy_age_12_19.dat}; 

    \addplot[fill={rgb:red,66;green,106;blue,127}, only marks,draw opacity=0,mark size=1.2pt] table [x=x, y=y] {data/accuracy_age_19_59.dat}; 

    \addplot[fill={rgb:red,66;green,106;blue,127}, only marks,draw opacity=0,mark size=1.2pt] table [x=x, y=y] {data/accuracy_age_59_116.dat};

    \addplot[fill={rgb:red,66;green,106;blue,127}, only marks,draw opacity=0,mark size=1.2pt] table [x=x, y=y] {data/accuracy_asian_age_12_19.dat}; 

    \addplot[fill={rgb:red,66;green,106;blue,127}, only marks,draw opacity=0,mark size=1.2pt] table [x=x, y=y] {data/accuracy_asian_age_59_116.dat}; 
    
    \end{axis}
\end{tikzpicture}

%  4741, 1995, 889, 709, 814, 218, 3322, 533, 19, 52
% 12 --- 443
% 15 --- 363
    \caption{Distribution of fairness multiplicity (i.e., group accuracy) across different intersections of racial and age groups in the UTKFace dataset. Each distribution is a condensed representation of a multiplicity sheet, containing the group accuracy of 65 independently trained models across all axes of multiplicity described in Section \ref{sec:setup}. Different groups have varying ranges of multiplicity, with a specially amplified variance from intersectional groups, highlighting the concerns and opportunities of multiplicity in fairness. Note: The minor perturbations along the x-axis for any group are only present for enhanced visualization and do not convey any additional signal.}
    \label{fig:fairness_scatter_utkface}
\end{figure*}

\subsection{Experiment Setup}
\label{sec:setup}

% We first provide an overview of the experiment setup, before jumping into the multiplicity analysis.

\paragraph{Dataset}
We will be studying the UTKFace dataset, containing facial images that have been labelled according to their perceived gender, race, and age. We will focus on the binary classification task of perceived gender. We split the dataset into $80\%$ training and $20\%$ testing, and we maintain the same split throughout our paper, i.e., we do not consider potential multiplicity introduced by the train-test splits.

\paragraph{Training Details}
By default, we train our models using a learning rate of 0.1, a batch size of 128, the data augmentation RandAugment \cite{cubuk2020randaugment}, the SGD optimizer, and the ResNet-18 architecture \cite{he2016deep}. For a simpler analysis, all models are trained from scratch, i.e., without the use of pre-trained weights.
We use a single random seed to control all forms of randomness in model training. We leave the decoupled analysis of various sources of randomness for future work. Finally, all models are trained with cross-entropy (CE) loss for 50 epochs, without any early stopping.

\paragraph{Axes of Multiplicity}
We will investigate the following different axes of multiplicity, (i) \textit{Learning Rate: } $\{0.1, 0.05, 0.01\}$, (ii) \textit{Batch Size: } $\{128, 256, 640\}$, (iii) \textit{Data Augmentation: } RandAugment \cite{cubuk2020randaugment} and TrivialAugment \cite{muller2021trivialaugment}, (iv) \textit{Optimizer:} SGD and Adam, (v) \textit{Model Architecture:} ResNet-18 \cite{he2016deep}, ResNet-50 \cite{he2016deep}, and WideResNet-50x2 \cite{zagoruyko2016wide}. 
We also compare multiplicity across changing randomness in model training.
% Additional results for a wider range of architecture choices are provided in Appendix \ref{sec:app_utkface_arch}.

\subsection{Group Fairness}

\begin{figure*}
    \centering
    \input{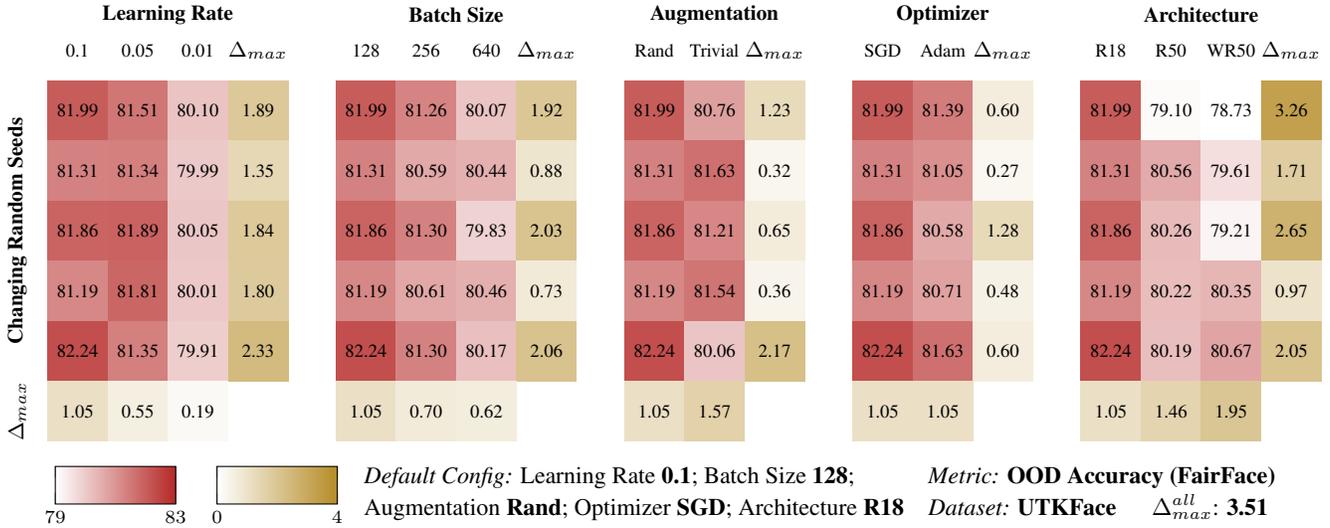}
    \caption{\textit{Multiplicity sheet} for \textbf{OOD Accuracy (FairFace)} on \textbf{UTKFace} dataset. R18/50: ResNet-18/50; WR50: WideResNet-50x2. The learning rate and batch size both seem to noticeably influence the OOD robustness, while it's the architecture choice that dominates any other factor for robustness multiplicity. The overall multiplicity ($\Delta_{max}^{all}$) is slightly higher than the fairness multiplicity seen previously.}
    \label{fig:robustness_multiplicity_utkface}
\end{figure*}

Group fairness is a measure of performance disparity between different protected groups, rooted in concerns of algorithmic bias propagated from the dataset to the model \cite{crawford2013hidden,barocas2016big,zhao2017men,abbasi2019fairness}. Traditionally, group fairness is measured as the difference in performance between different groups in the dataset. For accuracy under intervention in our setup, we calculate the accuracy on the minority group (which can also be extended to other groups). More specifically, we consider racial labels for fairness and measure the accuracy of the racial minority in the UTKFace dataset,  i.e., \textit{Asians}.

In Fig. \ref{fig:fairness_multiplicity_utkface}, we present the multiplicity sheet for Group Accuracy (Asian). We use the sheet to highlight the importance of random seeds in fairness and also contrast it to the multiplicity sheet for Accuracy in Fig. \ref{fig:accuracy_multiplicity_utkface}. As can be seen clearly from the $\Delta_{max}$ values of changing random seeds compared to different hyperparameter choices, random seeds have the most significant impact on fairness multiplicity. Moreover, we can also observe that the overall fairness multiplicity ($\Delta_{max}^{all}=3.24$) is three times higher than the accuracy multiplicity ($\Delta_{max}^{all}=1.12$). These trends of fairness variance and the impact of random seed have been previously noted in literature \cite{ganesh2023impact,soares2022your}, however here we show the ease with which they can be spotted in our multiplicity sheets.

We repeat the experiment for various groups and plot the distribution of fairness multiplicity across all axes of multiplicity, in Fig. \ref{fig:fairness_scatter_utkface}, with groups formed at the intersection of two different protected attributes, i.e., race and age. Our findings reveal that the severity of fairness multiplicity is even higher for intersectional groups. To put this into perspective, consider selecting a model from the distribution of models in Fig. \ref{fig:fairness_scatter_utkface}. While the choice may only affect the overall accuracy in the range of $92.05\%$ to $93.17\%$, it can significantly alter the accuracy for older Asian individuals, ranging from $88.46\%$ to $98.08\%$. To sum up, our analysis clearly shows the alarmingly high fairness variance, and the need to address this multiplicity such that deep learning models treat diverse groups fairly during deployment.

\subsection{Out-of-Distribution Robustness}

Out-of-distribution (OOD) robustness refers to the ability of a machine learning model to perform well on data points that are different from those it was trained on. Models that lack OOD robustness might make unreliable or incorrect predictions when faced with new, unfamiliar data, potentially leading to undesirable outcomes after deployment. Traditionally, OOD robustness is measured as the model's performance on an OOD dataset. Since it is already an accuracy metric, we do not perform any additional intervention for robustness. More specifically, we simply use the model's accuracy on the FairFace dataset \cite{karkkainenfairface}, a facial image dataset with a different distribution than UTKFace, as the measure of OOD robustness multiplicity.

In Fig. \ref{fig:robustness_multiplicity_utkface}, we present the multiplicity sheet for Accuracy on the FairFace dataset. We see the impact of learning rate, batch size, and architecture on robustness emerge from the multiplicity sheets, an unsurprising result based on existing work on the benefits of smaller batch size, larger learning rate, and higher risks of overfitting in bigger models \cite{jastrzkebski2017three,masters2018revisiting}. The range of overall robustness multiplicity ($\Delta_{max}^{all}=3.51$) is of the same range as fairness multiplicity, i.e. three times higher than accuracy multiplicity. Thus, addressing multiplicity in OOD robustness is essential to making sure the model doesn't fail even under minor distribution shifts.

\begin{figure*}
    \centering
    \input{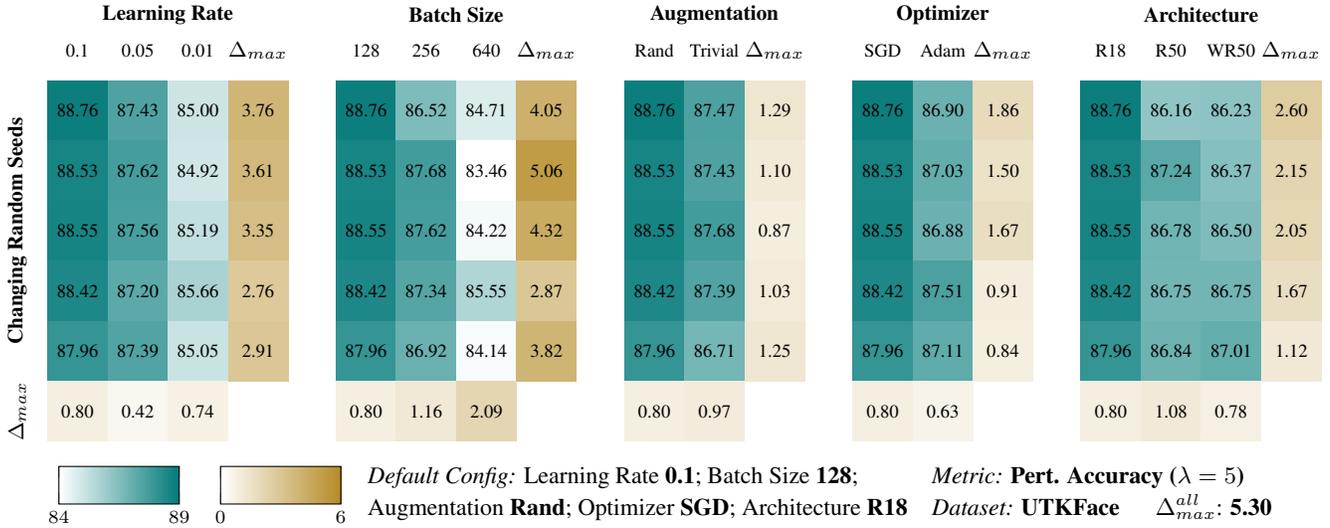}
    \caption{\textit{Multiplicity sheet} for \textbf{Perturbation Accuracy ($\lambda=5$)} on \textbf{UTKFace} dataset. R18/50: ResNet-18/50; WR50: WideResNet-50x2. The random seed has very little influence on the perturbation accuracy, while the default hyperparameter choices are noticably dominant over other hyperparameters, with the biggest drop caused by using a large batch size. The overall multiplicity ($\Delta_{max}^{all}$) is also quite high, five times larger than accuracy multiplicity, but clearly dependent on the choice of the rate parameter $\lambda$.}
    \label{fig:privacy_multiplicity_utkface}
\end{figure*}

\subsection{Differential Privacy}

Deep learning models tend to memorize data points from their training dataset, compromising the privacy of the individuals in the dataset. For instance, an adversary with access to only the outputs of the model is capable of extracting sensitive information from the model \cite{carlini2021extracting,shokri2017membership}. To address this issue, researchers often study differential privacy \cite{dwork2006differential}, which aims to make models trained on datasets differing at exactly one data point indistinguishable. One way to achieve this is by adding noise to the model's outputs. However, adding noise can also hurt the model's performance, thus creating a trade-off between privacy and accuracy.
It is this very trade-off that we will exploit to define our accuracy under intervention, i.e., we will measure the accuracy of the model under output perturbations from an exponential distribution with a fixed \textit{rate parameter} $\lambda$, for privacy multiplicity.

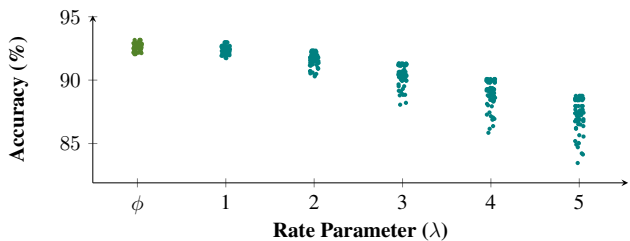
\begin{figure}[htbp]
    \centering
    \begin{tikzpicture}[scale=0.8]
    \begin{axis}[
        width=0.6\textwidth,
        height=0.25\textwidth,
        xmin=1,
        ymin=83, ymax=94,
        ylabel=\textbf{Accuracy (\%)},
        xlabel=\textbf{Rate Parameter ($\lambda$)},
        axis lines=left,
        enlarge x limits = true,
        enlarge y limits = true,
        xtick={1,2,3,4,5,6},
        xticklabels={$\phi$, 1, 2, 3, 4, 5},
        xticklabel style={align=center}
    ]

    \addplot[fill={rgb:red,78;green,123;blue,38}, only marks,draw opacity=0,mark size=1pt] table [x=x, y=y] {data/accuracy.dat};

    \addplot[fill={rgb:red,0;green,120;blue,120}, only marks,draw opacity=0,mark size=1pt] table [x=x, y=y] {data/accuracy_pert_1.dat};

    \addplot[fill={rgb:red,0;green,120;blue,120}, only marks,draw opacity=0,mark size=1pt] table [x=x, y=y] {data/accuracy_pert_2.dat}; 

    \addplot[fill={rgb:red,0;green,120;blue,120}, only marks,draw opacity=0,mark size=1pt] table [x=x, y=y] {data/accuracy_pert_3.dat}; 

    \addplot[fill={rgb:red,0;green,120;blue,120}, only marks,draw opacity=0,mark size=1pt] table [x=x, y=y] {data/accuracy_pert_4.dat};  

    \addplot[fill={rgb:red,0;green,120;blue,120}, only marks,draw opacity=0,mark size=1pt] table [x=x, y=y] {data/accuracy_pert_5.dat};
    
    \end{axis}
\end{tikzpicture}
    \caption{Distribution of privacy multiplicity (i.e., perturbation accuracy) across different values of the rate parameter $\lambda$. The higher the rate parameter, the larger the output perturbations, which in turn creates larger drops in accuracy and a larger range of multiplicity. Refer to Fig. \ref{fig:fairness_scatter_utkface} for further details on distribution visualization.}
    \label{fig:privacy_scatter_utkface}
\end{figure}

We present the multiplicity sheet for privacy by recording the Perturbation Accuracy with $\lambda=5$ in Fig. \ref{fig:privacy_multiplicity_utkface}. Interestingly, unlike other trustworthy metrics, the random seed has minimal impact on the privacy multiplicity. Instead, one hyperparameter choice along each axis is clearly the best when it comes to the privacy-accuracy trade-off, in line with the existing literature on practical tips for privacy~\cite{ponomareva2023dp}. Additionally, the overall privacy multiplicity range ($\Delta_{max}^{all}=5.30$) is almost five times larger than the accuracy multiplicity. These results are clearly dependent on the rate parameter $\lambda$ used to calculate accuracy under intervention and to emphasize this, we plot the distribution of privacy multiplicity for different rate parameter values in Fig. \ref{fig:privacy_scatter_utkface}. It is evident that choosing the right model by accounting for privacy multiplicity is crucial to achieving better privacy-utility trade-offs during inference, and this choice becomes even more critical with a decrease in privacy budget (i.e., higher values of rate parameter $\lambda$). 

\subsection{Security against Adversarial Attacks}

\begin{figure*}
    \centering
    \input{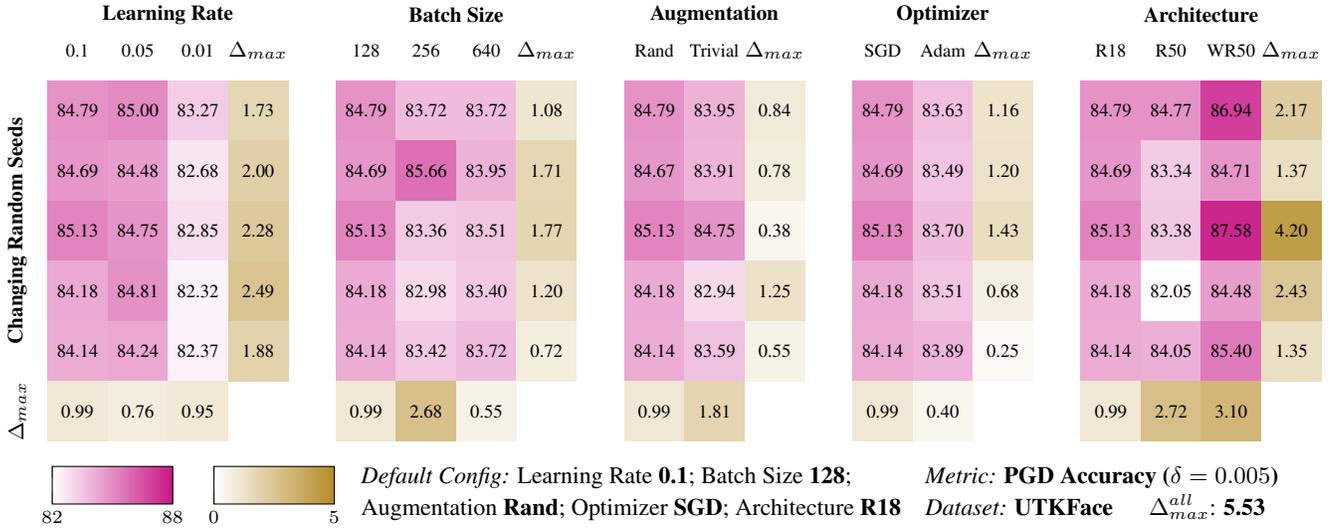}
    \caption{\textit{Multiplicity sheet} for \textbf{PGD Accuracy ($\delta=0.005$)} on \textbf{UTKFace} dataset. R18/50: ResNet-18/50; WR50: WideResNet-50x2. The architecture choice stands out as the most influential factor in security multiplicity. The overall multiplicity ($\Delta_{max}^{all}$) is five times larger than the accuracy multiplicity, dependent on the choice of the adversarial distance budget $\delta$.}
    \label{fig:security_multiplicity_utkface}
\end{figure*}

Machine learning models are vulnerable to various attacks that can manipulate the model's behaviour to suit the attacker's desires. One of the most common adversarial attacks studied in literature is the perturbation-based attack \cite{szegedy2014intriguing}, which takes advantage of the brittle decision boundaries of deep learning models. In this attack, the objective of the adversary is to perturb the input image by a minimum amount that can incite adversarial outputs, while keeping the perturbation imperceptible to the human eye. Instead of measuring the minimum distance of the perturbed image to the original image (measured as $L_\infty$), we will measure the accuracy under the intervention of a fixed distance budget represented by $\delta$. Specifically, we use projected gradient descent (PGD) \cite{madry2017towards} to progressively move out of the local minima until we reach the given distance budget, and then measure the accuracy of the model under these perturbations.

\begin{figure}[hbtp]
    \centering
    \begin{tikzpicture}[scale=0.8]
    \begin{axis}[
        width=0.6\textwidth,
        height=0.25\textwidth,
        xmin=1,
        ymin=65, ymax=94,
        ylabel=\textbf{Accuracy (\%)},
        xlabel=\textbf{Adversarial Distance Budget ($\delta$)},
        axis lines=left,
        enlarge x limits = true,
        enlarge y limits = true,
        xtick={1,2,3,4,5,6},
        xticklabels={0, 1e-4, 5e-4, 1e-3, 5e-3, 1e-2},
        xticklabel style={align=center}
    ]

    \addplot[fill={rgb:red,78;green,123;blue,38}, only marks,draw opacity=0,mark size=1pt] table [x=x, y=y] {data/accuracy.dat};

    \addplot[fill={rgb:red,199;green,21;blue,133}, only marks,draw opacity=0,mark size=1pt] table [x=x, y=y] {data/accuracy_pgd_1e_4.dat};

    \addplot[fill={rgb:red,199;green,21;blue,133}, only marks,draw opacity=0,mark size=1pt] table [x=x, y=y] {data/accuracy_pgd_5e_4.dat}; 

    \addplot[fill={rgb:red,199;green,21;blue,133}, only marks,draw opacity=0,mark size=1pt] table [x=x, y=y] {data/accuracy_pgd_1e_3.dat}; 

    \addplot[fill={rgb:red,199;green,21;blue,133}, only marks,draw opacity=0,mark size=1pt] table [x=x, y=y] {data/accuracy_pgd_5e_3.dat};  

    \addplot[fill={rgb:red,199;green,21;blue,133}, only marks,draw opacity=0,mark size=1pt] table [x=x, y=y] {data/accuracy_pgd_1e_2.dat};
    
    \end{axis}
\end{tikzpicture}
    \caption{Distribution of security multiplicity (i.e., accuracy after PGD) across different adversarial distance budget values $\delta$. A higher budget corresponds to a more powerful adversary, which in turn results in lower accuracy and higher security multiplicity. Refer to Fig. \ref{fig:fairness_scatter_utkface} for further details on distribution visualization.}
    \label{fig:security_scatter_utkface}
\end{figure}
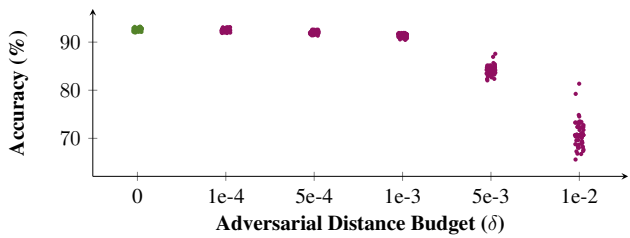

We present the multiplicity sheet for PGD Accuracy with $\delta=0.005$ in Fig. \ref{fig:security_multiplicity_utkface}. The trends for security multiplicity are similar to the accuracy multiplicity trends we observed previously, i.e., no single factor dominates the multiplicity, except for architecture choice. Surprisingly, the larger model ResNet-50 had a negative impact on security multiplicity, while the even larger but wider model WideResNet-50x2 improved it, which contradicts previous findings in literature \cite{wu2021wider} and raises interesting questions for future research. Similar to privacy multiplicity, the overall multiplicity range ($\Delta_{max}^{all}=5.53$) for security is almost five times larger than the accuracy multiplicity and depends on the adversarial distance budget $\delta$. We plot the distribution of security multiplicity for different adversarial distance budget values in Fig. \ref{fig:security_scatter_utkface}. Our results have shown that a model's robustness to adversarial attacks suffers from severe multiplicity, and needs to be addressed to provide robust models.

\section{Model Selection}

\begin{figure*}
    \centering
    \begin{tikzpicture}[scale=0.8]
    \begin{axis}[
        width=0.32\textwidth,
        height=0.25\textwidth,
        xmin=0.8,
        ymin=91, ymax=95,
        ylabel=\textbf{Accuracy (\%)},
        axis lines=left,
        enlarge x limits = true,
        enlarge y limits = true,
        xtick={1,2,3},
        xticklabels={Before, k=75\%, k=50\%},
        xticklabel style={align=center},
        title={\textbf{Group Accuracy (Age 59-116)}}
    ]

    \addplot[fill={rgb:red,120;green,120;blue,0}, only marks,draw opacity=0,mark size=1pt] table [x=x, y=y] {combined_data/accuracy_age_59_116_before.dat};

    \addplot[fill={rgb:red,120;green,120;blue,0}, only marks,draw opacity=0,mark size=1pt] table [x=x, y=y] {combined_data/accuracy_age_59_116_after75.dat};

    \addplot[fill={rgb:red,120;green,120;blue,0}, only marks,draw opacity=0,mark size=1pt] table [x=x, y=y] {combined_data/accuracy_age_59_116_after50.dat}; 
    
    \end{axis}
\end{tikzpicture}%
\begin{tikzpicture}[scale=0.8]
    \begin{axis}[
        width=0.32\textwidth,
        height=0.25\textwidth,
        xmin=0.8,
        ymin=50, ymax=73,
        axis lines=left,
        enlarge x limits = true,
        enlarge y limits = true,
        xtick={1,2,3},
        xticklabels={Before, k=75\%, k=50\%},
        xticklabel style={align=center},
        title={\textbf{OOD Accuracy (CelebA)}}
    ]

    \addplot[fill={rgb:red,120;green,120;blue,0}, only marks,draw opacity=0,mark size=1pt] table [x=x, y=y] {combined_data/accuracy_celeba_before.dat};

    \addplot[fill={rgb:red,120;green,120;blue,0}, only marks,draw opacity=0,mark size=1pt] table [x=x, y=y] {combined_data/accuracy_celeba_after75.dat};

    \addplot[fill={rgb:red,120;green,120;blue,0}, only marks,draw opacity=0,mark size=1pt] table [x=x, y=y] {combined_data/accuracy_celeba_after50.dat}; 
    
    \end{axis}
\end{tikzpicture}%
\begin{tikzpicture}[scale=0.8]
    \begin{axis}[
        width=0.32\textwidth,
        height=0.25\textwidth,
        xmin=0.8,
        ymin=58, ymax=90,
        axis lines=left,
        enlarge x limits = true,
        enlarge y limits = true,
        xtick={1,2,3},
        xticklabels={Before, k=75\%, k=50\%},
        xticklabel style={align=center},
        title={\textbf{Input Pert. Accuracy ($\lambda=1$)}}
    ]

    \addplot[fill={rgb:red,120;green,120;blue,0}, only marks,draw opacity=0,mark size=1pt] table [x=x, y=y] {combined_data/accuracy_pert_input_2_before.dat};

    \addplot[fill={rgb:red,120;green,120;blue,0}, only marks,draw opacity=0,mark size=1pt] table [x=x, y=y] {combined_data/accuracy_pert_input_2_after75.dat};

    \addplot[fill={rgb:red,120;green,120;blue,0}, only marks,draw opacity=0,mark size=1pt] table [x=x, y=y] {combined_data/accuracy_pert_input_2_after50.dat}; 
    
    \end{axis}
\end{tikzpicture}%
\begin{tikzpicture}[scale=0.8]
    \begin{axis}[
        width=0.32\textwidth,
        height=0.25\textwidth,
        xmin=0.8,
        ymin=65, ymax=81,
        axis lines=left,
        enlarge x limits = true,
        enlarge y limits = true,
        xtick={1,2,3},
        xticklabels={Before, k=75\%, k=50\%},
        xticklabel style={align=center},
        title={\textbf{PGD Accuracy ($\delta=0.01$)}}
    ]

    \addplot[fill={rgb:red,120;green,120;blue,0}, only marks,draw opacity=0,mark size=1pt] table [x=x, y=y] {combined_data/accuracy_pgd_1e_2_before.dat};

    \addplot[fill={rgb:red,120;green,120;blue,0}, only marks,draw opacity=0,mark size=1pt] table [x=x, y=y] {combined_data/accuracy_pgd_1e_2_after75.dat};

    \addplot[fill={rgb:red,120;green,120;blue,0}, only marks,draw opacity=0,mark size=1pt] table [x=x, y=y] {combined_data/accuracy_pgd_1e_2_after50.dat}; 
    
    \end{axis}
\end{tikzpicture}
    \caption{Distribution of multiplicity across unforeseen metrics under various degrees of model selection. The range of multiplicity across unforeseen metrics might remain unchanged even after we provide additional specifications for known trustworthy metrics, highlighting the severity of over-parameterization and the need to address multiplicity beyond just a checklist of metrics.}
    \label{fig:combined_scatter_utkface}
\end{figure*}
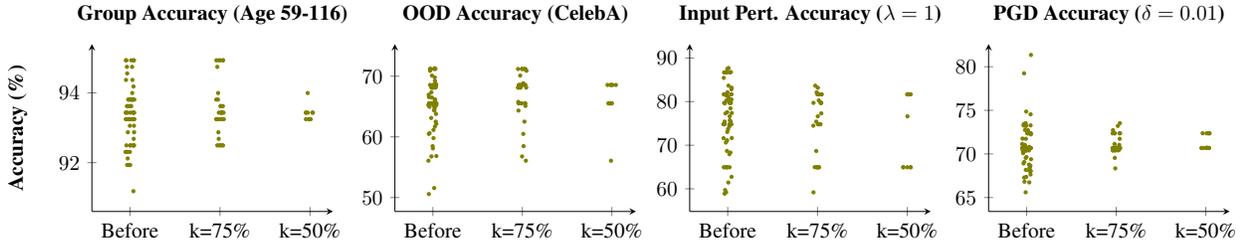

In our case study, we found significant multiplicity in various trustworthy metrics that can hurt model deployment, if left unchecked. To address this multiplicity, the literature suggests providing appropriate specifications during model selection \cite{black2022model}. This involves imposing additional constraints based on some chosen metrics, in our case the trustworthy metrics. For example, one can measure the fairness scores of the model under different hyperparameters and only choose the configurations with bias scores less than some threshold. This ensures that unfair models are not selected.

These recommendations stem from the belief that implementing extra measures during the selection of a model will decrease its variability, ensuring predictable behaviour upon deployment. However, as we will demonstrate in this section, over-parameterized models can still encounter unforeseen failure cases during deployment, which are not simply solved with appropriate specifications during model selection.

\paragraph{Model Specifications: } We first define the following criteria to simulate model selection. We choose models that rank in the top k\% of every metric under varying training configurations. We assess fairness by measuring accuracy for the Asian racial group, robustness by evaluating test performance on FairFace, privacy by measuring accuracy under output perturbations with a rate parameter $\lambda=5$, and security by measuring accuracy under PGD attacks with an adversarial distance budget of $\delta=0.005$. Our approach ensures that we only select models that meet the high standards for all four metrics mentioned above.

\paragraph{Unforeseen Circumstances: } We now introduce a new set of metrics to account for situations that were not previously considered in our specifications. To simplify the discussion, we will just make minor adjustments to the model specifications and create these 'unforeseen circumstances'. To test fairness, we will measure group accuracy for the age group $59-116$ instead of the Asian racial group. To test robustness, we will evaluate the performance on the CelebA dataset \cite{liu2015faceattributes} instead of FairFace. To test privacy, we will measure accuracy under input perturbations (with a rate parameter of $\lambda=1$) instead of output perturbations. Finally, to test security, we will increase the distance budget from $\delta=0.005$ to $\delta=0.01$, thus creating a stronger adversary.

In Fig. \ref{fig:combined_scatter_utkface}, we plot the distribution of multiplicity for all four unforeseen metrics before any model selection, and then after model selection for $k\%=75\%$ and $k\%=50\%$ respectively. We see a noticeable drop in unforeseen fairness and security multiplicities while maintaining decent fairness and security accuracy scores under intervention. However, we do not see this improvement in unforeseen robustness or privacy multiplicity. That is, despite the highly rigorous model selection on four different trustworthy ML metrics, the overall range of multiplicity in these two unforeseen metrics remains the same, and thus they will face the same issues during deployment. 
Clearly, incorporating additional specifications while selecting models can only provide limited assistance, leaving a substantial level of multiplicity that cannot be managed in the same way. Thus, addressing multiplicity with a checklist of trustworthy requirements is still likely to create models that face the same risks of failure in unforeseen circumstances, emphasizing the need for a more fundamental investigation into model multiplicity.
\section{Related Work}
\label{sec:related}

Model multiplicity has been an active subject of research in the deep learning literature, despite not being in the spotlight. Much of the related work in multiplicity is indirect, often disguised as research on the impact of hyperparameter choices or randomness on trustworthy ML~\cite{ganesh2023impact,soares2022your,jastrzkebski2017three,masters2018revisiting,ponomareva2023dp}.

Very few works in literature have focused solely on multiplicity. Black \etal\cite{black2022model} provides a discussion on the opportunities and concerns of multiplicity within the context of machine learning. However, their work is highly qualitative and does not provide any framework to quantify and measure multiplicity. On the other hand, D'Amour \etal\cite{d2022underspecification} offer a more quantitative perspective to underspecification in machine learning. Nevertheless, their analysis is fragmented across different case studies and does not provide a common language on multiplicity measurement that can be adapted for future works on model multiplicity.
\section{Conclusion and Future Work}

In this paper, we contribute to the discussion of model multiplicity, specifically in the context of image classification. By establishing a consistent and comprehensive language for multiplicity, we have created a foundation for  more effective communication in the field. Our multiplicity sheets offer an intuitive and structured approach to capturing the various facets of multiplicity. Furthermore, through a detailed case study, we demonstrated the practical implementation of our framework, shedding light on the complexities that arise when dealing with model multiplicity. The insights derived from the case study not only showcased the utility of our approach but also unveiled intriguing trends within the multiplicity scores. Finally, we show empirically that the challenge of model multiplicity cannot be simply resolved by providing additional specifications or constraints.

While we emphasize a specific structure for multiplicity sheets in this paper, it is important to acknowledge that further research is required to develop more effective methods for recording multiplicity. Moreover, our recommendation to use accuracy under intervention is primarily applicable to classification tasks. Nevertheless, the challenge of model multiplicity is a major issue in deep learning that goes beyond classification alone. Consequently, it is imperative that the community engages in further discussion on the topic of model multiplicity. We must shift away from treating multiplicity as an auxiliary discussion and bring it to the forefront to address potential unforeseen failures in real-world deployment scenarios and create truly trustworthy systems.

%% \etal~\cite{} or just ~\cite{}
%% Use \cref{} and \Cref{} instead of just ref

{\small
\bibliographystyle{ieee_fullname}
\bibliography{references}
}

\newpage
\appendix
\section{Image Classification on CIFAR10-Skewed}
\label{sec:app_cifar10}

The UTKFace dataset is an excellent choice for our main paper as it contains valuable metadata and has been extensively studied in trustworthy ML literature. However, it does come with certain limitations, such as its focus on binary classification tasks, and being confined to a highly specialized domain, i.e., facial images. To address these limitations, we will conduct a second case study using a skewed version of the CIFAR-10 dataset (more details below).

\subsection{Experiment Setup}
\label{sec:app_setup}

\paragraph{Dataset}
We will adopt the CIFAR10-Skewed setup from Wang~\etal\cite{wang2020towards} for our case study. In this setup, the 10 object classes of CIFAR10~\cite{krizhevsky2009learning} are divided into two groups, i.e., colour majority and grayscale majority. The first 5 classes (\textit{airplane, automobile, bird, cat, deer}) are marked as the colour majority, i.e., $95\%$ of images from these classes are left as is, while the other $5\%$ are converted into grayscale images. Conversely, the last 5 classes (\textit{dog, frog, horse, ship, truck}). are marked as the grayscale majority, i.e., $95\%$ of images from these classes are converted into grayscale images, while the other $5\%$ are left as is.

\paragraph{Training Details}
The training details are the same as the UTKFace Setup, except that we train the models for only 20 epochs on CIFAR10-Skewed.

\paragraph{Axes of Multiplicity}
The axes of multiplicity are the same as the UTKFace Setup, except for the model architecture. We will only use a modified ResNet18 model (adapted for CIFAR10 images of size 32x32) and we will not study multiplicity across changing architecture in this case study.

\subsection{Accuracy}

\begin{figure*}
    \centering
    \input{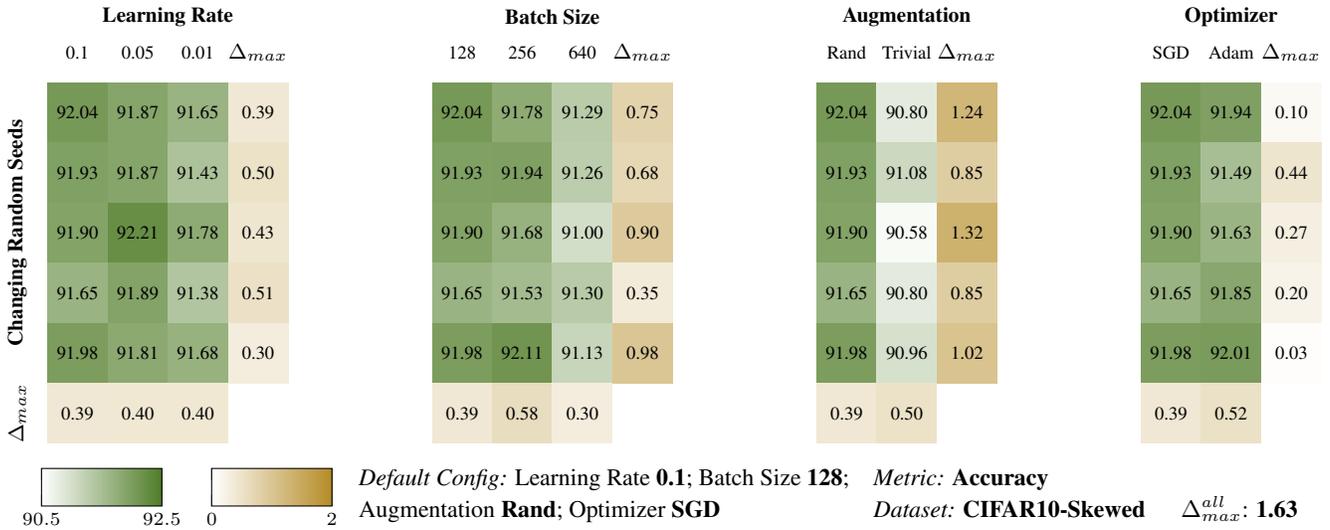}
    \caption{\textit{Multiplicity sheet} for \textbf{Accuracy} on \textbf{CIFAR10-Skewed} dataset.}
    \label{fig:accuracy_multiplicity_cifar10s}
\end{figure*}

The multiplicity sheet for accuracy on the CIFAR10-Skewed dataset is created in Fig. \ref{fig:accuracy_multiplicity_cifar10s}. Note that the test dataset for CIFAR10-Skewed, i.e., the dataset on which we measure this accuracy, is also skewed and has the same formulation as the training dataset defined above. The trends of accuracy multiplicity are quite similar to that of UTKFace, i.e., no significant accuracy variance is present across any hyperparameter choice or random seeds.

\subsection{Group Fairness}

\begin{figure*}
    \centering
    \input{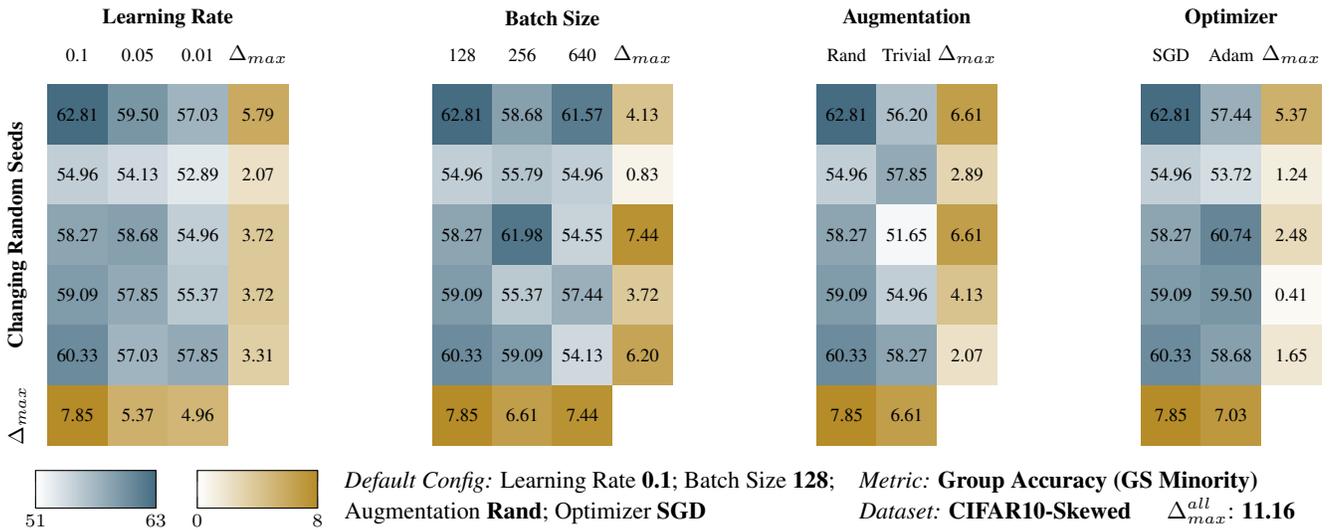}
    \caption{\textit{Multiplicity sheet} for \textbf{Group Accuracy (GS Minority)} on \textbf{CIFAR10-Skewed} dataset.}
    \label{fig:fairness_multiplicity_cifar10s}
\end{figure*}

For the CIFAR10-Skewed dataset, grayscale images in the 'colour majority' object classes and colour images in the 'grayscale majority' object classes are both minority groups. For this particular case study, we will measure the group accuracy of the GS Minority, i.e., the grayscale minority in the colour majority object classes, as the fairness score under intervention. The results are collected in Fig. \ref{fig:fairness_multiplicity_cifar10s}.

As previously noted, the multiplicity sheet in Fig. \ref{fig:fairness_multiplicity_cifar10s} highlights the importance of random seeds in fairness. While other hyperparameter choices do have a noticeable impact (in order - training data augmentation, batch size, learning rate, and the optimization algorithm), clearly the most consistently dominant source of fairness multiplicity is the randomness in model training. Moreover, the overall fairness multiplicity ($\Delta_{max}^{all}=11.16$) is almost seven times higher than the accuracy multiplicity ($\Delta_{max}^{all}=1.63$), further highlighting the severe impact of multiplicity on trustworthy metrics.

\subsection{Out-of-Distribution Robustness}

\begin{figure*}
    \centering
    \input{figure_scripts/appendix/robustness_multiplicity_cifar10s}
    \caption{\textit{Multiplicity sheet} for \textbf{OOD Accuracy (CIFAR10-GS)} on \textbf{CIFAR10-Skewed} dataset.}
    \label{fig:robustness_multiplicity_cifar10s}
\end{figure*}

We will use accuracy on a grayscale version of the CIFAR10 dataset \cite{krizhevsky2009learning} as the measure of our OOD robustness multiplicity. In Fig. \ref{fig:robustness_multiplicity_cifar10s}, we present the multiplicity sheet for Accuracy on the CIFAR10-GS dataset. The results show similar trends to robustness multiplicity for UTKFace, with both hyperparameter choices and the random seed being equally important in affecting the model's robustness. The range of overall robustness multiplicity ($\Delta_{max}^{all}=4.01$) is a little more than two times higher than accuracy multiplicity, which is unsurprising since despite being grayscale, the test images still belong to CIFAR10. A more severe robustness check on a dataset that is quite different from CIFAR10 might introduce and even higher OOD robustness multiplicity.

\subsection{Differential Privacy}

\begin{figure*}
    \centering
    \input{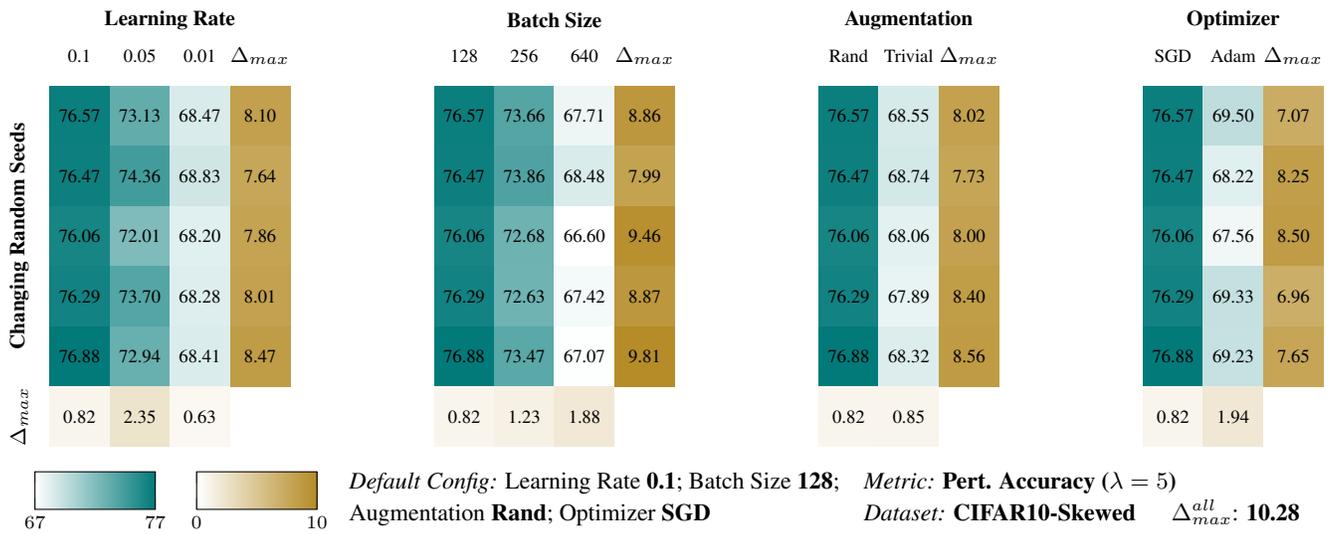}
    \caption{\textit{Multiplicity sheet} for \textbf{Perturbation Accuracy ($\lambda=5$)} on \textbf{CIFAR10-Skewed} dataset.}
    \label{fig:privacy_multiplicity_cifar10s}
\end{figure*}

We use the same perturbation and trade-off setup for privacy multiplicity as done for UTKFace, i.e., we will measure the accuracy of the model under output perturbations from an exponential distribution with a fixed \textit{rate parameter} $\lambda$. We present the multiplicity sheet for privacy by recording the Perturbation Accuracy with $\lambda=5$ in Fig. \ref{fig:privacy_multiplicity_cifar10s}. The same trends as Fig. \ref{fig:privacy_multiplicity_utkface} are noticed, i.e., the random seed has minimal impact and it's the hyperparameters that dramatically influence the privacy multiplicity, in line with the existing literature on practical tips for privacy~\cite{ponomareva2023dp}. The overall privacy multiplicity range ($\Delta_{max}^{all}=10.28$) is also almost six times larger than the accuracy multiplicity but would depend on the rate parameter $\lambda$.

\subsection{Security against Adversarial Attacks}

\begin{figure*}
    \centering
    \input{figure_scripts/appendix/security_multiplicity_cifar10s}
    \caption{\textit{Multiplicity sheet} for \textbf{PGD Accuracy ($\delta=0.005$)} on \textbf{CIFAR10-Skewed} dataset.}
    \label{fig:security_multiplicity_cifar10s}
\end{figure*}

We will use the same setup for security multiplicity as UTKFace, i.e., we use projected gradient descent (PGD) \cite{madry2017towards} to progressively move out of the local minima until we reach the given distance budget, and then measure the accuracy of the model under these perturbations. We present the multiplicity sheet for PGD Accuracy with $\delta=0.005$ in Fig. \ref{fig:security_multiplicity_cifar10s}. The trends are again similar to the ones seen in the main paper, i.e., no single factor dominates the multiplicity. The overall multiplicity range ($\Delta_{max}^{all}=4.51$) for security is almost five times larger than the accuracy multiplicity and clearly depends on the adversarial distance budget $\delta$.

\subsection{Discussion}

We have provided an additional case study on the CIFAR10-Skewed dataset as a companion to our main case study on the UTKFace dataset. These results help us cement certain trends, for example, the impact of random seeds on fairness, the impact of hyperparameter choices on privacy-utility trade-off, etc., all of which are unsurprising as these trends have been noted previously in the literature (albeit in isolated settings). We believe these experiments will serve as a useful companion to our main paper, and help establish the importance of multiplicity sheets in image classification.

\end{document}